\documentclass[letterpaper, 10 pt, conference]{ieeeconf}  
\IEEEoverridecommandlockouts      
\usepackage{times}
\usepackage[pdftex]{graphicx}
\usepackage{graphics}
\usepackage[cmex10]{amsmath}
\usepackage{url}
\usepackage[caption=false,font=footnotesize]{subfig} 
\usepackage{array}
\usepackage{tabularx}
\usepackage{multirow}
\usepackage{multicol}
\usepackage{booktabs}

\usepackage{amssymb}
\usepackage{amsmath}
\usepackage[all]{xy}    
\usepackage{ifthen}
\usepackage[usenames,dvipsnames]{color}
\usepackage{enumerate}

\makeatletter
\let\NAT@parse\undefined
\makeatother
\usepackage[numbers]{natbib}

\newcommand{\secref}[1]{Section~\ref{#1}}

\newcommand{\figref}[1]{Figure~\ref{#1}}

\newcommand{\myparagraph}[1]{\vspace{0.1in}\noindent\textbf{#1}}

\newboolean{draft-mode}
\setboolean{draft-mode}{true}
\newcommand{\sidenote}[1]{\ifthenelse{\boolean{draft-mode}}{\marginpar{\tiny\raggedright\textsf{\hspace{0pt}#1}}}{}}
\DeclareRobustCommand{\arnote}[1]{\ifthenelse{\boolean{draft-mode}}{\textcolor{blue}{\textbf{AR:#1}}}{}} 
\DeclareRobustCommand{\nfnote}[1]{\ifthenelse{\boolean{draft-mode}}{\textcolor{red}{\textbf{NF:#1}}}{}} 
\DeclareRobustCommand{\pynote}[1]{\ifthenelse{\boolean{draft-mode}}{\textcolor{LimeGreen}{\textbf{PY:#1}}}{}} 
\DeclareRobustCommand{\ncdnote}[1]{\ifthenelse{\boolean{draft-mode}}{\textcolor{green}{\textbf{NCD:#1}}}{}} 

\newcommand*{\Cdot}{\raisebox{-0.25ex}{\scalebox{1.75}{$\cdot$}}}

\title{\LARGE \bf A Summary of Team MIT's Approach to the Amazon
  Picking Challenge 2015}

\author{\authorblockN{Kuan-Ting Yu$^1$, Nima Fazeli$^2$, Nikhil Chavan-Dafle$^2$, 
Orion Taylor$^2$, Elliott Donlon$^2$, \\
Guillermo Diaz Lankenau$^2$, and Alberto Rodriguez$^2$}
\authorblockA{ $^1$Computer Science and Artificial Intelligence
  Laboratory ---
  Massachusetts Institute of Technology \\
  $^2$Mechanical Engineering Department ---
  Massachusetts Institute of Technology\\
  {\tt\small <peterkty, nfazeli, nikhilcd, tayloro, edonlon, diazlank,
    albertor>@mit.edu}} \thanks{We would like to thank the support of
  ABB Robotics and Amazon Robotics for organizing the challenge.}}
  
\begin{document}
\maketitle
\thispagestyle{empty}
\pagestyle{empty}

\begin{abstract}
  The Amazon Picking Challenge (APC)~\citep{apcwebsite}, held alongside
  the International Conference on Robotics and Automation in May 2015
  in Seattle, challenged roboticists from academia and industry to
  demonstrate fully automated solutions to the problem of picking
  objects from shelves in a warehouse fulfillment scenario.
  Packing density, object variability, speed, and reliability are the
  main complexities of the task. The picking challenge serves both as
  a motivation and an instrument to focus research efforts on a specific manipulation problem.
  In this document, we describe Team MIT's approach to the
  competition, including design considerations, contributions, and
  performance, and we compile the lessons learned. We also describe
  what we think are the main remaining challenges.
\end{abstract}



\section{Introduction}
\label{sec:introduction}

From the introduction of the pallet jack in 1918 to the irruption of
KIVA systems in 2003, warehouse automation has seen a wealth of
advancements in the last century.
%
%
The story is one of removing unstructuredness in material handling, in
part by intelligent and efficient packing or palletizing. To this day,
the greatest limiting factor is in the ability to handle, with speed
and reliability, individual objects presented with non-trivial packing
density. Of particular interest are the tasks of picking and
re-stocking objects from shelves, bins, or boxes, where grasping
remains an unsolved problem.

The economic value of robust and flexible object picking and
re-stocking in warehouses is large and hard to quantify.  Half a
billion new pallets are made every year, and Amazon alone sold 426
items per second in the peak of 2013 Christmas season, all picked
and boxed by hand.

\begin{figure}
\centering
\includegraphics[width=3.4in]{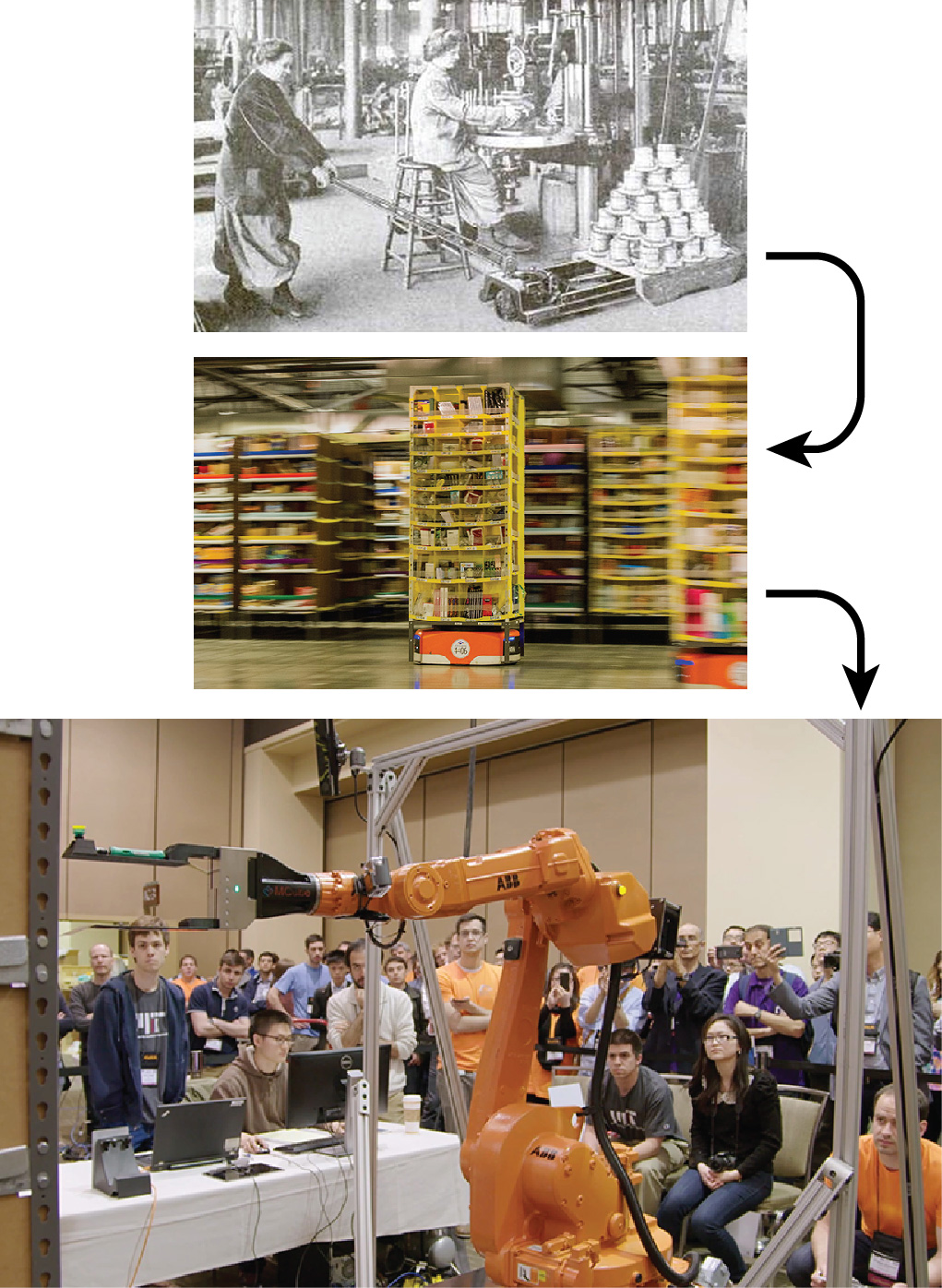}
\caption{(top) Pallet-jack from 1918. ``This new truck gets under the
  load and lifts it!''~\cite{pallet_jack}. (center) Kiva robot moving a
  pod in a warehouse. (bottom) Team MIT's robot picking from a pod in
  the 2015 Amazon Picking Challenge.}
\label{fig:warehouse_automation}
\end{figure}

The Amazon Picking Challenge (APC)~\citep{apcwebsite}, organized by
Amazon Robotics, and advised by expert academics in the field of
robotic manipulation, aims at focusing research efforts at addressing
this very timely and potentially transformative technology of
shelf-picking and self-restocking.

The 2015 challenge, described in detail in \secref{sec:problem}, is a
first step in that direction, and tasked participating teams with the
development of a fully automated system to locate and pick a small set
of objects cluttered by other objects and the surrounding shelving
structure. This paper describes Team MIT's approach, who finished 
second among 30+ competitors.



\section{The 2015 Amazon Picking Challenge}
\label{sec:problem}
The 2015 Amazon Picking Challenge posed a simplified version of the
general picking and re-stocking problem. Teams were given a space of
2x2 meters in front of a shelving unit lightly populated with objects,
and 20 minutes to autonomously pick as many items as possible from a
list of desired items.

Objects ranged in size, shape and material, and their arrangement was
adequately simplified for a first challenge of this kind. The teams
were provided with a list of 25 items prior to the competition. The
exact 12 items that were to be picked from a shelf were unknown until 2 minutes
before the competition and their poses and configurations were only
discovered by the robot on run time.

\subsection{The 25 Items}

\begin{figure}
  \begin{center}
	\includegraphics[height=3.1in]{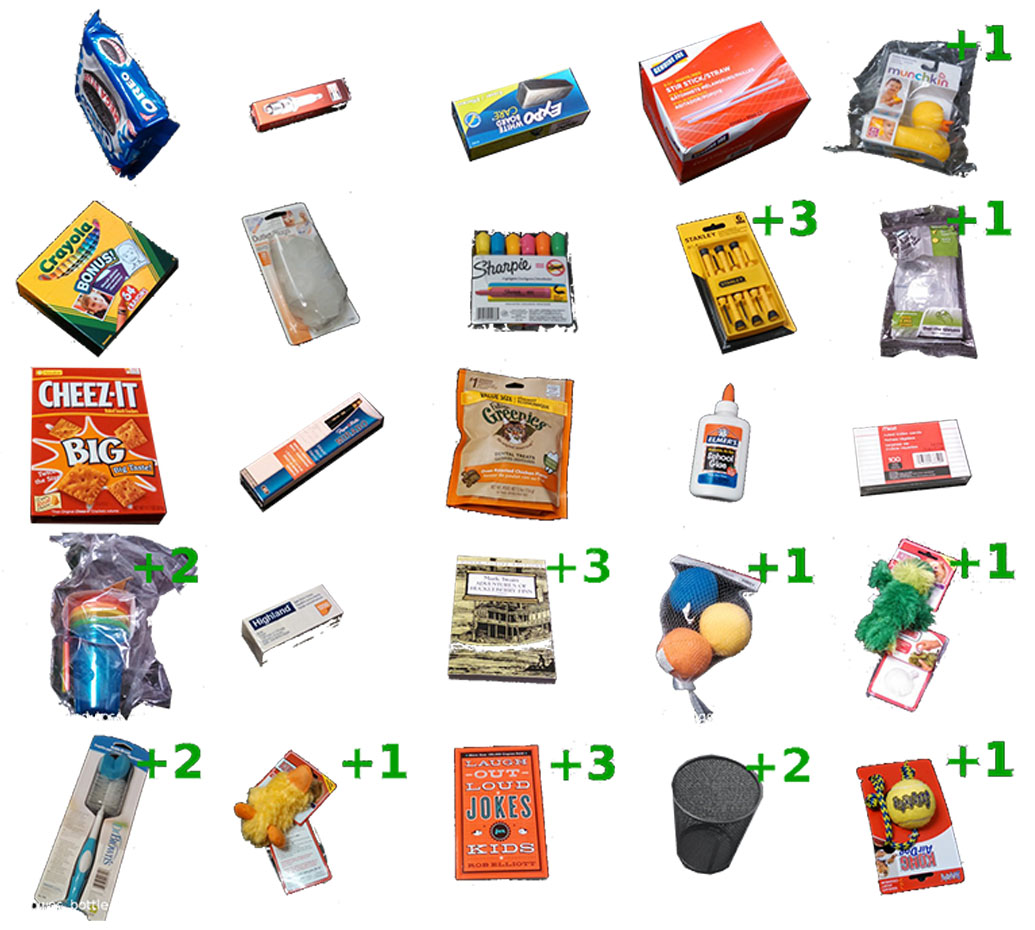}
  \end{center}
	\caption{The 25 items in the competition. From left to right
          and top to bottom: Oreo box, boxed sparkplug, boxed
          whiteboard eraser, large box of straws, plastic wrapped
          duck, crayola box, plastic box of outlet plugs, plastic set
          of highlight markers, set of small screwdrivers, plastic
          wrapped glasses, large Cheez-it box, pencil box, bag of cat
          food, glue bottle, plastic wrapped note cards, plastic
          wrapped set of baby cups, plastic wrapped sticky notes,
          book, set of 3 foam balls, furry frog, packaged bottle
          brush, furry duck, book, meshed metallic pencil cup, and
          packaged tennis ball.}
	\label{fig:items}
\end{figure}

The objects used in the competition are depicted in
\figref{fig:items}. According to the challenge organizers, these were
selected to span a wide range of sizes, shapes and materials, and are
representative of a significant portion of the item transactions that
are handled on a daily basis in an Amazon warehouse. The items were
chosen with the intent to provide the participants with realistic
challenges such as:

\begin{itemize}
\item[$\Cdot$] \textbf{Large sized objects.} Items that may not fit
  into regular gripper spans or that may collide with the shelf during
  the extraction process, such as large boxes and books.
\item[$\Cdot$] \textbf{Small objects.} Items whose size require high
  location accuracy for picking, but that are small for perception to
  recover accurate models.
\item[$\Cdot$] \textbf{Packaging.} The reflective
  packaging on some of the items complicates perception, especially for depth sensors that
  rely on time of flight or optical triangulation.
\item[$\Cdot$] \textbf{Deformable shapes.} Some items are non-rigid,
  which requires perception and manipulation techniques that can cope
  with the variability of their shapes and compliance.
\end{itemize}

\subsection{The Shelf}

\begin{figure}
  \begin{center}
	\includegraphics[height=3.1in]{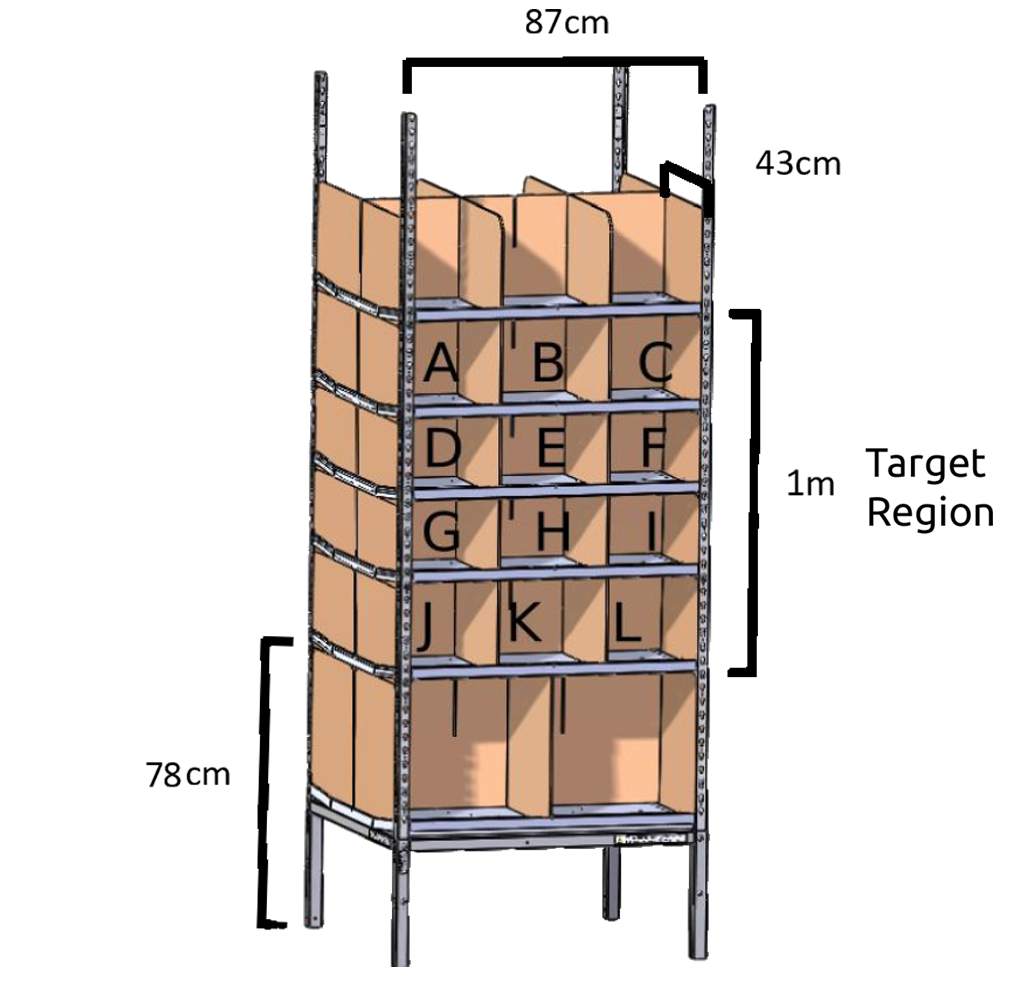}
  \end{center}
  \caption{The competition shelf. The target region defined for the
    competition is a cuboid of 1m tall by .87m wide and
    .43m deep, and composed of 12 individual bins of slightly
    different dimensions.}
	\label{fig:shelf}
\end{figure}

The competition shelf is a standard Kiva pod used in Amazon
warehouses, a rigid structure comprised of individual bins. To reduce
the reachability requirements, the competition is constrained to the
12 bins in the center, as shown in \figref{fig:shelf}, which defines a
cuboid of 1 meter high by 87 cm wide by 43 cm deep. 

\begin{figure}
  \begin{center}
	\includegraphics[width=3.2in]{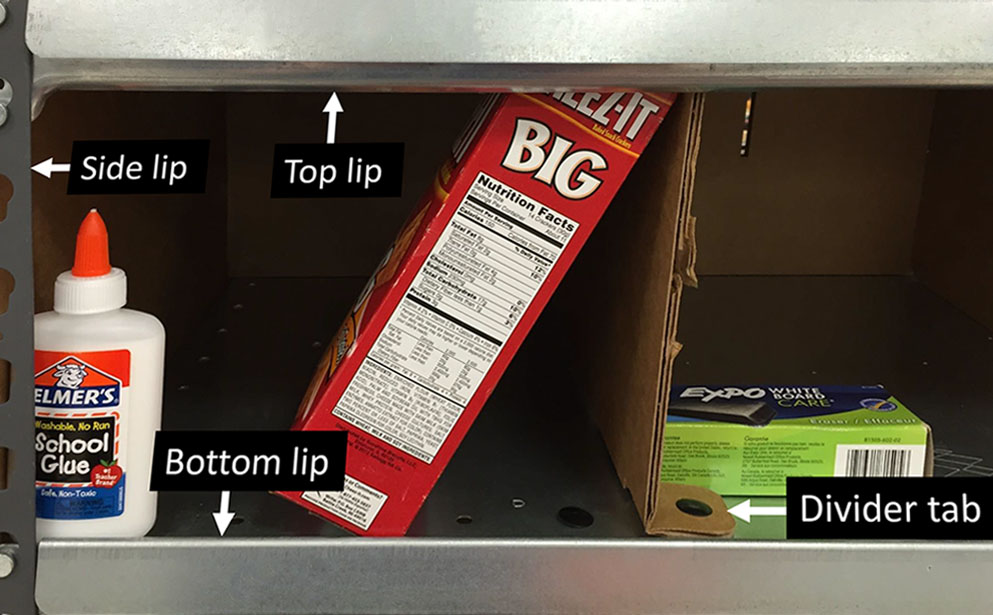}
  \end{center}
	\caption{Shelf bin lips and divider tab.}
	\label{fig:shelf_detail}
\end{figure}

The structure has tolerances, asymmetries, and construction artifacts
that deviate it from a perfect array of walls and shelves. These
are the most relevant ones:
\begin{itemize}
\item[$\Cdot$] The walls and shelves are not equi-distributed. This
  introduces differences in the nominal size of the openings of each
  individual bin, with height ranging between 19 and 22cm, and width
  between 25 and 30 cm.
\item[$\Cdot$] Each bin has a lip on the bottom and top edges, as
  shown in \figref{fig:shelf_detail}, which impedes exposing an object
  by sliding it.
\item[$\Cdot$] The lateral bins have a lip of the exterior edge, as
  shown in \figref{fig:shelf_detail}, which impedes exposing an object
  by pulling on it.
\end{itemize}
The approach we describe here is based on planning of accurate end
effector trajectories, which requires a detailed understanding of
these deviations.

Finally, also worth noting, is the metallic bottom of the structure,
which produced bad reflections from depth sensors and proved to be an
impediment for accurate estimation of the location of the shelf by
model fitting to pointcloud data. 

\subsection{Scoring}
The performance of a team was evaluated according to:
\begin{itemize}
\item[$\Cdot$] Picking a target item from a bin with one item: 10
  points.
\item[$\Cdot$] Picking a target item from a bin with two items: 15
  points.
\item[$\Cdot$] Picking a target item from a bin with three items or
  more: 20 points.
\item[$\Cdot$] Damaging any item induced a 5 point penalty.
\item[$\Cdot$] Dropping a target item from more than 30cm incurred a 3
  point penalty and a non-target item incurred in 12 point penalty.
\item[$\Cdot$] Some items had bonus points because of their
  increased difficulty.
\end{itemize}

\subsection{Simplifications}
It is worth noting that this first instance of the competition had
many simplifications from the general picking and re-stocking
problem, most of them to the object arrangement, and facilitated
perception and frontal or top picking:
\begin{enumerate}
\item[$\Cdot$] Items were placed close to the front of the bins,
  effectively reducing the necessary  workspace of the robots.
\item[$\Cdot$] Items were arranged next to each other, but not on
  top or behind each other.
\item[$\Cdot$] Items were not tightly packed.
\item[$\Cdot$] Items were not repeated inside individual bins.
\item[$\Cdot$] The time allowed (20 minutes) was much larger than the
  average time that a human would take to solve the task ($\sim$1 minute).
\end{enumerate}


\section{Design Philosophy}
\label{sec:rationale}

Our core design philosophy was driven both by the motivation to win
the competition, as well as by looking towards a scalable solution
that could eventually tackle the larger problem of picking and
re-stocking in a real scenario. From the very beginning, we believed
on the importance of aiming towards a system capable of picking a
tightly packed set of varied objects, with speed and reliability,
which drove most of our decisions.

For this competition, our design approach was based on: 1) The
accuracy, controllability and reachability of a large and stable
industrial robotic arm; 2) The flexibility and thin profile of custom
made flat fingers, and 3) The versatility of a set of highly developed
primitive actions.

Also critical was our effort towards an integrated system that did not
need assembly or wiring in the competition. Our platform was
self-contained except for a master external PC. Our
system was supported by a heavy base that contained
the robot, cameras, PCs for image processing, power adapter,
compressor, UPS, and all the necessary cabling. We believe the
reduction in the need for integration in the competition was
instrumental to the reliability of the system. 

The following sections describe in detail the hardware and software
subsystems.


\section{Hardware System Overview}
\label{sec:hardware}

The hardware infrastructure is based on an industrial ABB 1600ID robot
arm coupled with a parallel-jaw gripper with custom-designed fingers,
and integrated cameras. The particular robot model was chosen for its
sufficient workspace, position accuracy, and high speed (we used a
tool max speed of 1 m/s for the TCP during the competition). The robot
has a hollow wrist and purpose-built canals that allow routing cables
and airlines from the base to the gripper, which is important to
maximize maneuverability and preventing cables and connectors from
being pulled through the interaction with objects and
environment. This section describes in detail the gripper, cameras,
and other feedback sensors integrated in the system.

\subsection{End-Effector}
\label{sec:ee}
\begin{figure}
  \begin{center}
    \includegraphics[width=3.3in]{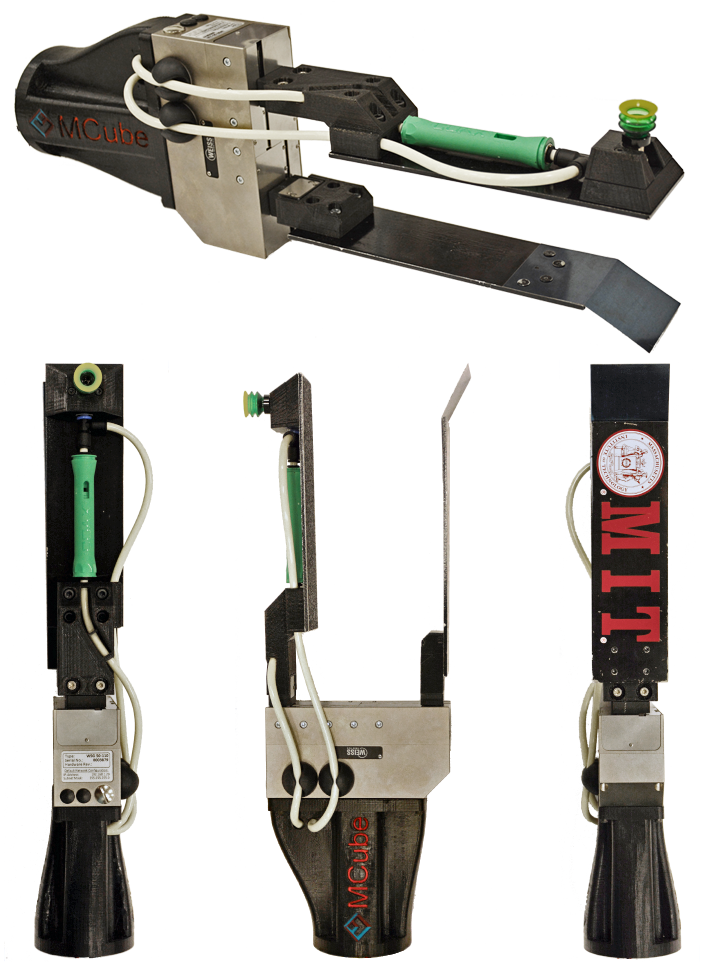}
  \end{center}
  \caption{A parallel-jaw gripper with custom designed fingers with
    suction system and compliant spatula.}
  \label{fig:endeffector}
\end{figure}

The design of the end-effector was driven by the desire to integrate
in a compact solution three different picking modalities:
\emph{grasping}, \emph{suction}, and \emph{scooping}. As shown in
\figref{fig:endeffector}, these take place by the combined action of a
parallel-jaw gripper with flat fingers, a suction cup, and a compliant
spatula.

\myparagraph{Gripper Selection.}
The gripper of choice was the parallel-jaw gripper WSG 50 from german
manufacturer Weiss Robotics, with the following features:
\begin{itemize}
\item[$\Cdot$] \emph{Opening range of 110mm.} Sufficient for most
  objects in at least two of their main axes. 
  Unfortunately, the max opening of the gripper governs the
  nominal size of the gripper's body, which we wanted to make sure
  could fit inside the bin both in vertical and horizontal directions.
\item[$\Cdot$] \emph{Max gripping force of 70N.} With
  sufficient friction, this gives enough gripping force to maneuver
  all objects without risk of dropping.
\item[$\Cdot$] \emph{Position and Force control.} Both were
  instrumental. Uncertainties in the exact location
  of the walls and bottom required opening and closing with force
  control to achieve a desired preload on the spatula.
\end{itemize} 
Finally, we would like to highlight that we did not find significant
advantages in available three fingered grippers due to their
relatively large size.

\myparagraph{Gripper Customization.}
The custom designed fingers shown in \figref{fig:endeffector} are
comprised of high-strength aluminum plates, which provides
stiffness, durability, and low mass. 

One finger incorporates a hardened spring steel spatula, designed to
deform elastically when pre-loaded against the shelf's walls or
bottom, which is key for inserting fingers under or on the side of
objects flushed against the wall. Additionally, there is a suction
system on the top finger for lifting items that are difficult to
grasp.

The second finger has an integrated suction cup with an in-line
vacuum-Venturi device. A vacuum sensor that gives feedback
on strength of seal with the object and the valve to regulate the air flow are hosted in
the 3D printed wrist.

\subsection{Cameras}
\label{sec:visonsensors}
Robust perception is a key component of the challenge. Distinguishing
the identity of objects, and recovering an accurate pose is essential
for a robust execution. The nature of the challenge poses significant
challenges, including occlusions, tight view angles, and noise due to
bad reflections.

\begin{figure}
  \begin{center}
	\includegraphics[width=3.40in]{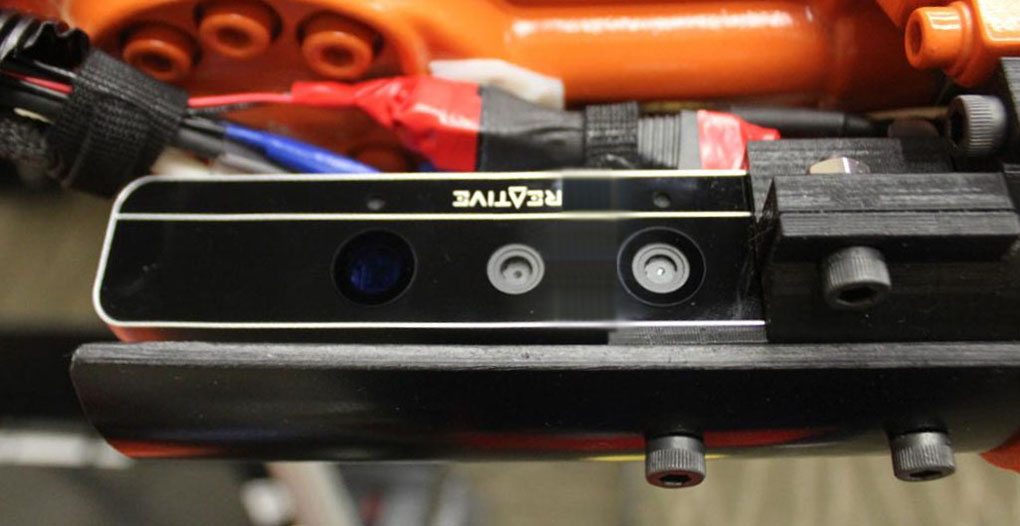}
  \end{center}
  \caption{Intel RealSense camera mounted on the 5th axis of the robot.}
	\label{fig:realsense}
\end{figure}

We manually optimized the selection of camera and camera location to
maximize the view angles for the target objects; avoid occlusions from
robot arm, gripper, and shelf; avoid collisions with objects and
environment; and safety of the device. For the sake of robustness, our
system combined the use of:
\begin{itemize}
\item[$\Cdot$] Two Kinect2 cameras mounted to the base of the robot at about
  1.7m from the ground, providing two well calibrated top-side views
  of the bin. Kinect2 uses time-of-flight to measure depth, which
  allows multiple cameras to operate simultaneously. It senses in the
  $\left[0.8-4\right]$m range, with a resolution of 512x424. It has
  problems with plastic packaging due to multiple reflections.
\item[$\Cdot$] An arm-mounted Intel RealSense camera, as seen in
  \figref{fig:realsense} provides close-up, accurate depth
  point-clouds. RealSense uses structured-light for recovering depth.
  It senses in the $\left[0.2-1.2\right]$m range, and has a resolution
  of 640x480.
\end{itemize}

\subsection{Computers}
\label{sec:computers}
Our control architecture is composed of:
\begin{itemize}
\item[$\Cdot$] A master computer (Lenovo ThinkStation, Intel Xeon
  E3-1241 CPU, 32 GB RAM, Nvidia Titan X GPU), that controls and
  supervises the entire system. The GPU in this system is used to run
  the perception system which is further discussed in
  \secref{sec:perception}.
\item[$\Cdot$] Two auxiliary compact computers (Gigabyte Brix from
  Intel i7-4770R CPU, 16 GB RAM) integrated in the robot platform, capture, pre-process and filter pointclouds, to
  reduce the load on the master computer.
\end{itemize}
Also worth noting that, due to the problems with the max length of the
USB connection of the RealSense camera, one of the compact computers
had to be mounted on axis 3 of the robot. Ethernet connection seems to
be preferable for its robustness and flexibility.

\begin{figure*}
  \begin{center}
    \includegraphics[width=6.0in]{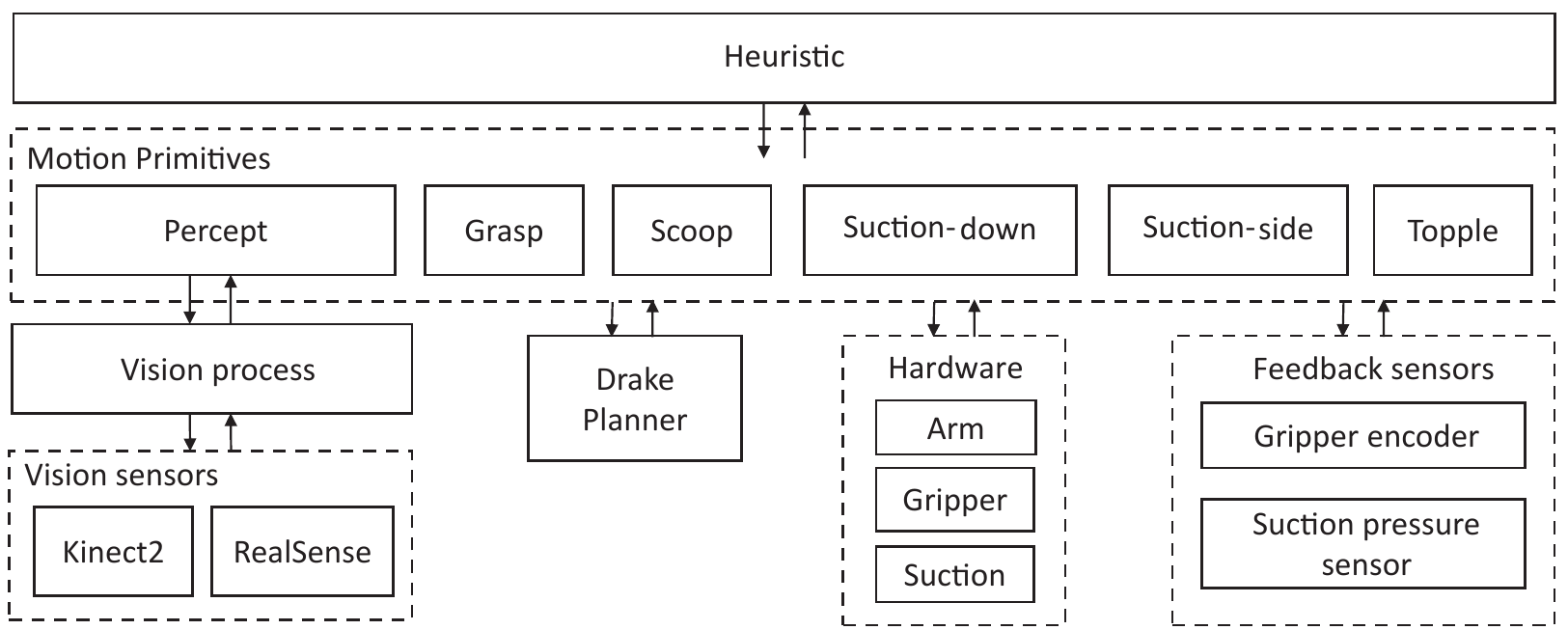}
  \end{center}
  \caption{System overview. A main heuristic decides what motion
    primitive to executed based on sensor input and system state.}
  \label{fig:system}
\end{figure*}

\section{Software System Overview}
\label{sec:software}

At a high level, the autonomy of our system is governed by a series of
primitive actions---involving both action and perception---as well as a
centralized heuristic, acting as the brain of the system, that decides
what primitive to execute based on the goal item to pick and feedback
from cameras and sensors in the end-effector (\figref{fig:system}).
The individual primitives take care of the fine details of the
interaction with the environment, low level planning, and execution,
also based on feedback from cameras and sensors in the end-effector.

The software architecture is based on the ROS framework~\citep{ros},
overseeing the interaction between perception, motion primitives,
planning, and user interface, which we describe in detail in this section.

\subsection{Perception}
\label{sec:perception}
The role of perception is to find the identity and location of all the
objects in a bin in the presence of occlusions, narrow view points
and noise due to bad reflections. This is a challenging problem, and
in our experience with trying different existing solutions, still a
relatively unsolved one. 

Our approach was driven by interaction with the software from Capsen
Robotics~\citep{capsen}, who gave us access to a GPU implementation of
an algorithm that fits existing models to depth-only pointclouds. Our
final approach was:
\begin{itemize}
\item[1.] Filter points outside the convex hull of the shelf, and bad
  reflections returned as NaN. To ease the filtering, we transform the
  pointclouds into shelf frame, where walls are axis aligned. We then store
  the information in a point-based mask.
\item[2.]  We specify physical constraints for object detection inside
  a given bin. Objects must lie within the bin's walls and on the
  floor. To that end, the resulting object pose should intersect a
  up-shifted and thickened version of the bottom of the bin.
\item[3.] We feed the pointcloud, mask, constraints, and the ID of the
  objects in the bin to the Capsen Robotics
  software~\citep{capsen}. This outputs an array of scene hypotheses
  associated with a score of log likelihood, each containing an array
  of item IDs and 6D poses.
\item[4.] We reject hypotheses with the center of mass outside the bin, since Capsen treats constraints as
  soft-constraints.
\item[5.] We select the hypothesis with the highest score. In case of
  multiple instances of the target item, we randomly pick one of them
  as the target object.
\item[6.] For the sake of robustness, we run this algorithm
  on depth images from the closest Kinect 5 times, and repeat another 5 times on depth
  images from the RealSense camera.
\end{itemize}
We would like to note that, although our final approach was based only
on depth information, we believe now the combination of depth
and RGB to be essential for a robust perception solution.


\subsection{Motion Primitives}
\label{sec:primitives}
One of the main philosophical choices in our approach is the use of
motion primitives rather than a straight motion planning approach. A
motion primitive is an action defined by its goal of achieving a
specific type of manipulation. The main reasons are the need for
robustness and speed. By constraining the full capability of robot
motion to a small set of predefined families of motions, we can
focus on high-end performance for specific task, giving us a better
understanding of the expected performance of the system our system while also speeding up planning since the plan (except for low
level joint motions) is pre-computed during the development of the
primitive. We proceed with a description of the primitives we
implemented.

\begin{figure*}
  \includegraphics[width=2.30in]{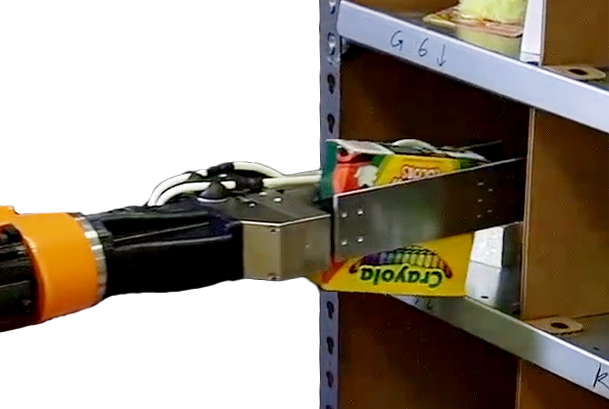}
  \includegraphics[width=2.30in]{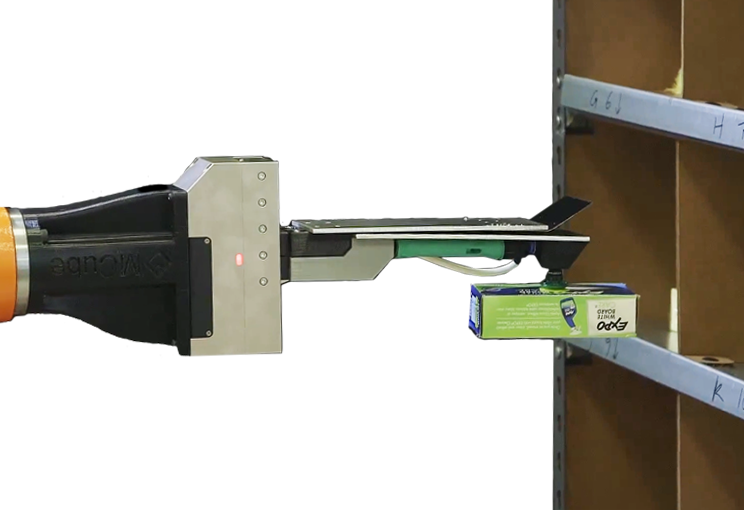}
  \includegraphics[width=2.30in]{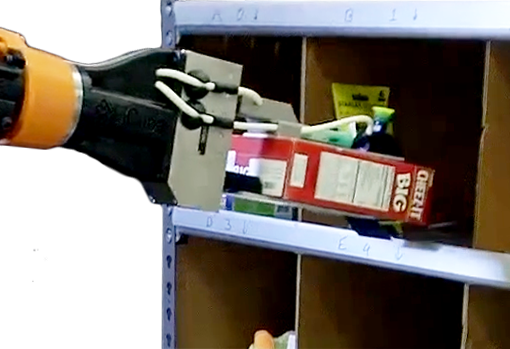}
  \caption{Examples of primitives: (left) grasp, (center) suction, and
    (right) scoop.}
  \label{fig:primitives}
\end{figure*}

\myparagraph{Percept.}
Responsible for moving the robot arm to place the arm-mounted camera
in front of the desired bin, and getting out of the way from the fixed
cameras. The primitive is also in charge of turning on and off the
cameras to prevent IR light to cross-affect cameras. It is also worth noting
that Kinect2 tends to blind RealSense because of the high-intensity IR strobe.

For the arm-mounted camera, we pre-select two advantageous viewpoints
for each bin, so no arm motion planning is required on real time.
Of special interest is the difficulty of having the camera mounted on
the 5th axis instead of at the end-effector. This prevents the camera
from colliding with objects and environment, but reduces is workspace.
Planning arm motions is done by searching for an IK solution with a
tolerance, rather than specifying a hard constraint on the 6DoF pose
of the camera.

For most items, and for the sake for robustness, we use both Kinect
and RealSense. The different technologies used by both cameras, give
us a small gain by running both. For large, non-reflective objects,
however, Kinect2 is sufficient.


\myparagraph{Grasp.}
%
%
%
This primitive aims to end up with a vertical parallel jaw grasp of
the target item in between the two flat fingers.
It relies on known geometric information of the objects and the
maximum opening of the gripper, which we use to define a set of
advantageous grasp, a well as their expected finger opening. Based on
the object shape and pose, the primitive decides: the
 trajectory of approach of the gripper, its pose before closing the
fingers, and the expected finger opening after the grasp.

Rather than searching over the space of all possible grasping
trajectories, our algorithm is based on the following principles:
\begin{itemize}
\item[$\Cdot$] The nominal strategy is to approach the object straight
  from the front of the shelf towards the detected pose of the object.
\item[$\Cdot$] Search over the pitch and yaw orientation of the gripper in its
  approach to the object, to avoid collisions of the gripper with the
  shelf (including the multiple lips).
\item[$\Cdot$] If the target item is too close to a wall, we make use of the
  spatula finger to add pretension on the wall so that it can slide on
  the side of the object. Note that by employing the compliant
  spatula, and exploiting the environment, object configurations that
  are very difficult to reach for traditional gripping systems, turn
  into very robust and easy to reach grasps.
\end{itemize}


\myparagraph{Scoop.}
Primitive specially designed to make use of the spatula at the end of one
of the fingers, as shown in \figref{fig:endeffector}.
The combination of the compliance of the spatula with the force
control of the opening of the gripper, allows the robot to flush one
finger against the floor of the bin with a controlled preload, and
objects to be ``scooped'' between the fingers by pushing them against
the back of the shelf. 

Scoop is a powerful and very robust primitive that, again, makes use
of pushing~\citep{Mason1986, Dogar2010} and the environment to reduce
uncertainty~\citep{Eppner2015} and extend
dexterity~\citep{ChavanDafle2014,ChavanDafle2015a}, which is specially
apt for objects difficult to perceive or grasp because of being small,
deformable, or flat.

\myparagraph{Suction.}
Basic, but very functional primitive useful for objects with exposed
flat surfaces. We limit our implementation to objects with horizontal
or vertical surfaces.

For example, in the case of a horizontal-down suction, the process
simply involves lowering the suction cup down onto the exposed face.
Whether or not the target item presents a suctionable face depends
upon its shape and current pose. The process is as:


\begin{itemize}
\item[1.] The robot positions the suction cup above the centroid of
  the object and lowers the cup. The compliance of the cup and the
  force control of the gripper tells the robot when to stop the down
  motion.
\item[2.] We check whether the height at which the motion was blocked
  matches the expected height (from geometric object models).
\item[3.] The suction cup lifts, and the system checks the pressure
  sensor to determine whether suction was successful or not. We make a
  max of 5 attempts to suck the object with small variations in the
  X-Y plane.
\end{itemize}

Suction is specially advantageous for objects that are flat and wide.  However, our choice of suction cup could not form a seal with
bags or furry objects, which renders it un-usable for those
objects. In general, picking up an object with suction is either very
easy or impossible, with very few cases in between.

\myparagraph{Topple.} 
This is a helper motion primitive whose goal is not to
pick, but to change the configuration of an object that cannot be
picked by the other primitives. It is called in particular when the
exposed face of an object is tall and wide.  
In that case, suction is not possible because the gripper would collide with
the top part of the bin, the opening of the gripper might not be wide
enough for grasping or scooping.
In this scenario it may be possible to topple
the object into a new configuration that will be grasp-able, suction-able
or scoop-able. 

Topping is implemented by pushing the object with a relatively rapid
motion above its center of mass, so that the object rotates above its
back supporting edge. The question of whether an object can be toppled
can be answered beforehand~\citep{Lynch1999}, but deciding whether
the primitive is working as expected in real time is still a difficult
problem. Our work-around is call perception again after executing
toppling.

\myparagraph{Push-Rotate.}
Push-Rotate is another helper function that aims at turning an object
to expose a graspable side.
The maximum gripper opening limits the set of graspable objects and
poses. For most of the objects in this iteration of the challenge, one
or more of the dimensions of the dimensions was graspable with our
current gripper. The combination of push-rotate and toppling allow us
to deal with most corner-cases.

Push-Rotate chooses where and how to push the object to make the smaller 
dimension face the shelf front and generates the robot trajectory accordingly.
After execution, as in the case of toppling, perception is called
again to asses the new pose of the object.

\subsection{Motion Planning}
\label{sec:planning}

Each primitive is specified as a series of 6DOF end-effector poses and
gripper openings through which the robot will need to sequence. 
To execute them we use the inverse kinematics planner from the Drake
package developed by the Robotic Locomotion Group at
MIT~\citep{drake}, which provides a detailed sequence of joint
trajectories for the robot to follow. 
Given that the series of end-effector poses are already very
descriptive of the intended motion and are designed with
collision avoidance in mind, we do not take into consideration
collisions with the shelf when generating the trajectories themselves.

\subsection{Heuristic}
\label{sec:Heuristics}

At the task level, the autonomy of the system is driven by a
heuristic, composed of a state machine and a prioritized list of
primitive actions to try first for different object configurations.

The heuristic is in charge of processing the work order and sorting
target objects by difficulty (based on the desired object and amount
of clutter). It then goes one by one through target items from easiest
to most difficult. For each target:
\begin{itemize}
\item[1.] It scans the associated bin and decides the type of pose of the
  object;
\item[2.] Chooses and executes a strategy (set of primitives chained together)
  based on the specific object, its pose type, and its prioritized
  list of primitives.
\item[3.] Evaluates success, and based on the outcome, and the number
  of attempts, decides whether to reattempt with a different strategy
  or skip the item.
\end{itemize}
The choice of strategy is a crucial task. In our implementation, the
specific priorities given to each primitive and object pose were fine-tuned with the help of numerous experiments to estimate success rates
based on few object configurations. It would be desirable to develop
a more systematic approach to optimize the heuristic used to decide
what is the best primitive to use for a given bin configuration and
desired item.

\subsection{Calibration}
\label{sec:calibration}
Our approach relied heavily on an accurate calibration of the location
and geometry of the shelf, as well as the location of the cameras.
The calibration begins with locating the base of the robot centered
with the shelf and at an ideal distance from the shelf, optimized to
maximize the dexterous workspace. This location had a tolerance of
about 5cm.

We then used a semi-automated process, where the robot iterated
through a series of guarded moves to accurately locate the walls,
tops, bottoms, and lips of all bins in the shelf. A time consuming,
and ideally avoidable procedure, but necessary for the open loop
execution of many of the steps of the designed primitives.
Finally we calibrated the cameras with the help of the robot. The
Kinect camera point clouds are calibrated using
AprilTags~\citep{olson2011apriltag} attached to the robot end-effector.


\subsection{Networking}
\label{sec:networking}
The communication between different processes and machines was handled
through the Robot Operating System (ROS) architecture with
publish/subscribe structure when dropping messages was not critical,
and services for communications that require reliability.

The interface between the overall system and the Drake planner run in
MATLAB was through a standard TCP connection, and JSON format files as
marshalling method.
%
%

\subsection{User Interface}
\label{sec:ui}
Finally, although the system is meant to be autonomous, the user
interface is a key component specially during the developing phase.

We make use of \texttt{rviz} to visualize states of the system, and
facilitate input into the system, such as positions and orientations
with interactive markers, which is useful, for example to manually
approximate the calibration of shelf and cameras, and manually input
the object pose for testing purposes.
%
%
The visualization is also useful for pre-screening planned arm motions
which can prevent collisions due to coding errors. Our system
visualizes: 1) state of robot and gripper; 2) trajectory generated by
the planner; 3) pointclouds and object pose estimated from perception
algorithm, which give us clues as to what components of the software
architecture err when failures occur.  

The overall plan for handling an order is visualized with a dynamic
webpage that displays the order of items to pick, and current item.

Finally, for managing the execution and stopping all processes, we use
Procman~\citep{leonard2008perception}.  Our system runs about ten
simultaneous processes that are run in three different
machines. Automating the starting and stopping of the system saves a
lot of time, and avoids many errors that might happen during the
starting protocol. We found Procman particularly superior to the
standard \texttt{roslaunch} from ROS.



\section{At the Competition}
\label{sec:competition}
The competition was a motivating factor, especially to work on the
integration and robustness of the overall system. As already mentioned
before, we put a strong emphasis on minimizing the amount of
integration that would need to be done on the day of the
competition, which led us to use a heavy base that contained the
entire system, except for the master computer.

\myparagraph{Setup procedure.}
The robot was shipping as a whole unit, except for the outer structure
that was holding the cameras in place. Once in the venue, we assembled
the structure, and had to replace the compressor because it broke
during shipping. After the system was up and running, we followed the
calibration procedure and the system was ready to run.

During the design we also tried to minimize the time it would take to
replace broken components. For example we had a second gripper which
we could replace in just a few minutes.  We did by making sure that
all cables and tubes going to the gripper had a connector behind
the wrist. It turned out to be a useful feature, given that we actually
had to replace the gripper during the testing phase because a
miss-calibration and accidentally bending one of the fingers.

\myparagraph{The competition run.}
At the beginning of the 20-minute run, the gripper went into a reboot
mode, most likely due heavy network traffic in the TCP/IP communication even before making the first
attempt to pick the first item. We were penalized by 5 minutes, the
gripper got back to normal state, we restarted the execution, and
everything run more or less smoothly after that.

Our system picked 7 items out of 12 items, lightly damaging the
packaging of one of them, in about 7 minutes before the motion
supervision of the robot stopped the execution due to a torque
overload in one of the joints. The system could have easily continued
by pressing play on the robot controller, but the rules did not allow,
since that would have constituted a human intervention.
In total, we scored 88 points finishing second in the competition
among the 30+ international teams from both industry and academia.


\section{Discussion}
\label{sec:discussion}

The picking challenge is designed to be a multi-year effort where
solutions will need to keep pace with gradual increases in the
complexity of the challenge. We enjoyed the experience and hope to
participate again next year.
We believe the competition was particularly enriching because:
\begin{itemize}
\item The focus that comes from a real and well defined task. By
  formulating a specific problem it removes some of the academic
  freedom, but at the same time prevents many of the biases we tend to
  introduce in doing so, often in the direction of problems that we
  know how to solve or feel comfortable with.
\item The lack of constraints in the hardware or software was
  motivating and led to a wealth of approaches, almost as many as
  different participating teams. In future iterations teams might
  start converging in their approach based on previous experiences. We
  hope this document helps in understanding what we did and why we did
  it, as well as opinionating on what was instrumental and
  detrimental of our approach.
\end{itemize}

\subsection{Future Work}

\myparagraph{Perception.}
In our experience it is very difficult to solve this problem without
robust perception, and this will require a clever combination of depth
and RGB data. We have observed many situation where either one or the
other lack to provide enough information to positively locate an
object.
It is also important to notice the difficulty of plastic and
reflective packaging, where both color and depth information fail
often to recover useful information.
%

We believe an improvement to our current system might come from fusing
multiple views. The accuracy of the robot, jointly with existing
algorithms for pointcloud fusion, should give us more dense,
multi-facetted and reliable pointclouds of the scene, which can lead
to more reliable perception.

\myparagraph{Calibration.}
Our system relies on an accurate calibration of the geometry of the
shelf. That is not reflective of the real problem, so we would like to
step away from the need for accurate calibration, by relying more on
camera information and develop reactive versions of our primitives by
closing the loop with either force or tactile data.

\myparagraph{Thin manipulation.}
Probably the most exciting direction for the future for us is the
focus on thin manipulation, or manipulation in tight spaces. We would
like to be able to pick and re-stock objects in scenes with high
object density and in situations of tight packing. We finish with the
motivating example of picking a book from a shelf of many.





\bibliographystyle{IEEEtranN} 
{\footnotesize \bibliography{icra16-apc-sub1}}
\end{document}